\documentclass[sn-mathphys,Numbered]{sn-jnl}
\usepackage{graphicx}%
\usepackage{multirow}

\usepackage{amsthm}%
\usepackage{mathrsfs}%
\usepackage[title]{appendix}%
\usepackage{xcolor}%
\usepackage{textcomp}%
\usepackage{manyfoot}%
\usepackage{booktabs}
\usepackage{algorithm}%
\usepackage{algorithmicx}%
\usepackage{algpseudocode}%
\usepackage{listings}%
\usepackage{amsmath, amssymb, diagbox}
\usepackage{float}
\usepackage{xcolor} 

\usepackage{slashbox}
\usepackage{pifont}
\usepackage{lineno}
\usepackage{soul}

\begin{document}

\title[Automated vision-based assistance tools in bronchoscopy: stenosis severity estimation]{Automated vision-based assistance tools in bronchoscopy: stenosis severity estimation}

\author[1]{\fnm{Clara} \sur{Tomasini}}\email{ctomasini@unizar.es}
\author[1]{\fnm{Javier}\sur{Rodriguez-Puigvert}}
\author[2]{\fnm{Dinora}\sur{Polanco}}
\author[2]{\fnm{Manuel}\sur{Viñuales}}
\author[1]{\fnm{Luis} \sur{Riazuelo}}
\author[1]{\fnm{Ana C.} \sur{Murillo}}

\affil[1]{\orgname{DIIS, i3A. Universidad de Zaragoza}, \city{Zaragoza}, \country{Spain}}
\affil[2]{\orgname{Hospital Universitario Miguel Servet}, \city{Zaragoza}, \country{Spain}}

\abstract{
\textbf{Purpose:} Subglottic stenosis refers to the narrowing of the subglottis, the airway between the vocal cords and the trachea. Its severity is typically evaluated by estimating the percentage of obstructed airway. This estimation can be obtained from CT data or through visual inspection by experts exploring the region. However, visual inspections are inherently subjective, leading to less consistent and robust diagnoses. No public methods or datasets are currently available for automated evaluation of this condition from bronchoscopy video.
\\
\textbf{Methods:}
We propose a pipeline for automated subglottic stenosis severity estimation during the bronchoscopy exploration, without requiring the physician to traverse the stenosed region. Our approach exploits the physical effect of illumination decline in endoscopy to segment and track the lumen and obtain a 3D model of the airway. This 3D model is obtained from a single frame and is used to measure the airway narrowing.
\\
\textbf{Results:}
Our pipeline is the first to enable automated and robust subglottic stenosis severity measurement using bronchoscopy images. The results show consistency with ground-truth estimations from CT scans and expert estimations, and reliable repeatability across multiple estimations on the same patient. Our evaluation is performed on our new Subglottic Stenosis Dataset of real bronchoscopy procedures data.
\\
\textbf{Conclusion:} 
 We demonstrate how to automate evaluation of subglottic stenosis severity using only bronchoscopy. 
Our approach can assist with and shorten diagnosis and monitoring procedures, with automated and repeatable estimations and less exploration time, and save radiation exposure to patients as no CT is required. Additionally, we release the first public benchmark for subglottic stenosis severity assessment.}

\maketitle
\section{Introduction}
Subglottic Stenosis (SGS) is a condition characterized by the narrowing of the airway in the subglottis, just below the vocal cords and before the trachea, which results in airway obstruction airflow restriction.  
The percentage of narrowed airway or Stenosis Index (SI) is obtained by comparing the obstructed airway area to that of a posterior part of the airway without obstruction.
In clinical practice, physicians estimate the SI using either a CT scan, which involves exposure to radiation, or during a bronchoscopy procedure. In bronchoscopy, the estimation is performed visually or by fitting endotracheal tubes of known dimensions through the stenosis. Bronchoscopy-based estimates are inherently subjective, making diagnosis less robust and potentially leading to non-optimal treatment strategies~\cite{begnaud2015measuring}.

The subjectivity of bronchoscopy-based assessments has been previously shown ~\cite{murgu2013subjective,begnaud2015measuring}. 
When asked to assess stenosis grade in 1447 patients {\cite{murgu2013subjective}}, physicians tended to wrongly classify it (53\% of cases assessed the wrong severity level), mostly underestimating it (91\% of wrongly classified cases were underestimations). While CT offers a more objective alternative, its utility for continuous monitoring is limited due to the patient cumulative radiation exposure. 
To reduce assessment subjectivity, there is a need for a standardized measuring tool~\cite{begnaud2015measuring}, using bronchoscopy images to avoid repeated patient radiation exposure from CT scans. 
This work introduces the first pipeline to enable automated and robust estimation
of subglottic stenosis severity using bronchoscopy images only. 

\begin{figure}[tb]
    \centering
    \includegraphics[width=0.85\linewidth]{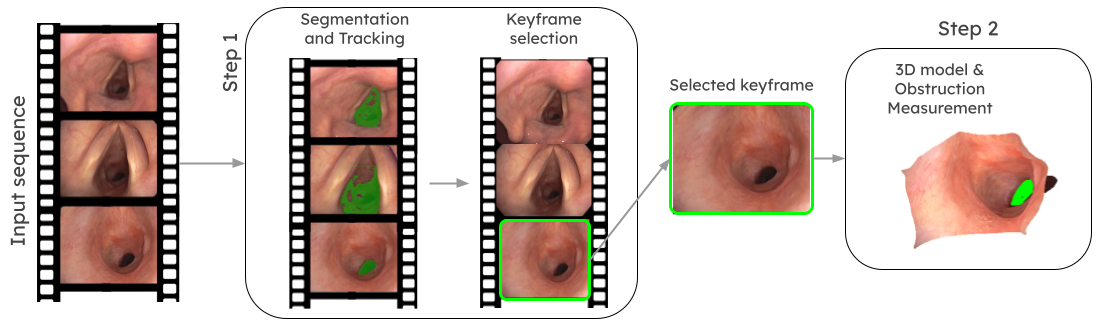}
    \caption{Overview of the proposed automatic stenosis estimation approach. Our two-step pipeline takes as input the bronchoscopy video, starting in front of the vocal cords. In each frame, the lumen is segmented and tracked until the camera 
    reaches the subglottis and starts to see the stenosis. At that selected keyframe, we compute a 3D reconstruction of the explored airway, in which the stenosis is measured.}
    \label{fig:intro}
\end{figure}

However, public real-world bronchoscopy image data is notably scarce, in particular including subglottic stenosis recordings. This motivates the self-supervised nature of the main step in our approach, as discussed later, and also the collection and release of our Subglottic Stenosis Dataset from real procedures.  
Figure~\ref{fig:intro} summarizes the proposed pipeline, that relies on illumination decline in endoscopy to segment and track the airway lumen, and to obtain a 3D
model of the explored airway 
applying LigthDepth~\cite{Rodriguez-Puigvert_2023_ICCV}. 
The 3D model is obtained from a single frame automatically selected right after the vocal cords, and is used to estimate airway narrowing. 

Overall, the contributions of this work are twofold: 
\begin{itemize}
    \item A novel approach to enable automated Subglottic Stenosis measurement based only on bronchoscopy video, acquired during regular physicians exploration but  without requiring the endoscope to traverse the stenosis. Our work is the first to demonstrate the use of \textit{illumination decline} as a supervisory signal for the acquisition of 3D reconstructions in bronchoscopy procedures. Our promising results indicate that our approach can help reduce the subjectivity involved in diagnosing of SGS. Besides, it can reduce exploration time, and minimize the use of CT scans, benefiting patients and assisting physicians in diagnosis and monitoring procedures.  

    \item  A novel benchmark, Subglottic Stenosis Dataset, for SGS severity estimation\footnote{Data available at \burl{https://sites.google.com/unizar.es/subglottic-stenosis-estimation/home}}, with data obtained from real bronchoscopy procedures along with ground-truth from CT scans and expert estimations. Public benchmarks are needed to encourage further research on automated stenosis evaluation.
\end{itemize}

\section{Related Work}

\textbf{Stenosis estimation.} Physicians generally agree~\cite{begnaud2015measuring} that a standardized stenosis measuring tool is needed to reduce uncertainty and variability observed in traditional clinical methods. 
Fully automatic approaches for SGS severity estimation have been proposed, though most use other imaging modalities such as CT, MRI or Optical Coherence Tomography~\cite{sharma2016quantitative}. 
Bronchoscopy images have been used in \cite{banach2023computer} to estimate tracheal stenosis. The narrowing is measured by considering two frames in inspiration and expiration phases and using depth estimation to evaluate the area of the opening without and with obstruction~\cite{banach2023computer}. 
Alternatively, airway areas can be estimated through segmentation of the obstructed lumen and reference healthy tracheal ring~\cite{sanchez2015toward}. Segmentation of the lumen can be obtained from depth estimation~\cite{keuth2023weakly} or image pixels intensity~\cite{sanchez2014line}, since the lumen usually corresponds to the deepest or darkest area of the image. However, these methods cannot be directly applied to SGS, as they rely on tracheal anatomical features. The subglottis, being the lower portion of the larynx cavity, does not exhibit the cartilage rings and the breathing deformation seen in the trachea and used in \cite{sanchez2015toward,banach2023computer}. As very little data is available for stenosis estimation, training-free methods such as SLIC superpixels~\cite{achanta2010slic}, and foundation models for image segmentation like SAM~\cite{kirillov2023segment} can also prove relevant. 
For specific SGS estimation using bronchoscopy images, \cite{murgu2009morphometric,francom2019clinical} propose a semi-automatic method that requires physician input to measure the stenosis using image processing software. \\
\textbf{SGS estimation data.} No publicly available datasets currently exist for stenosis estimation, limiting the development of a fully automatic tool. A few datasets exist however for the task of lumen segmentation, such as the CVC-LumenDB dataset~\cite{sanchez2014line}, which has very low resolution, or the Phantom dataset~\cite{visentini2017deep}, consisting of images obtained in a phantom airway instead of real bronchoscopy images. 
Endoscopy datasets with high resolution real images are more widely available for gastroscopy and colonoscopy, like EndoMapper~\cite{azagra2023endomapper} or Kvasir~\cite{pogorelov2017kvasir}.  

\section{Methodology}
\subsection{Overview}\label{sec:overview} 
During a bronchoscopy, physicians place the bronchoscope in front of the vocal cords to ensure a clear view of the airway cavity to navigate towards the trachea, and avoid visual occlusions that could affect accurate stenosis assessment. They proceed through the airway towards the lumen, which usually appears as the darkest area in the image, and visually estimate stenosis by evaluating the lumen visible in the image, using their prior knowledge and the bronchoscope diameter.

\begin{figure}[tb]
    \centering
    \includegraphics[width=0.75\linewidth]{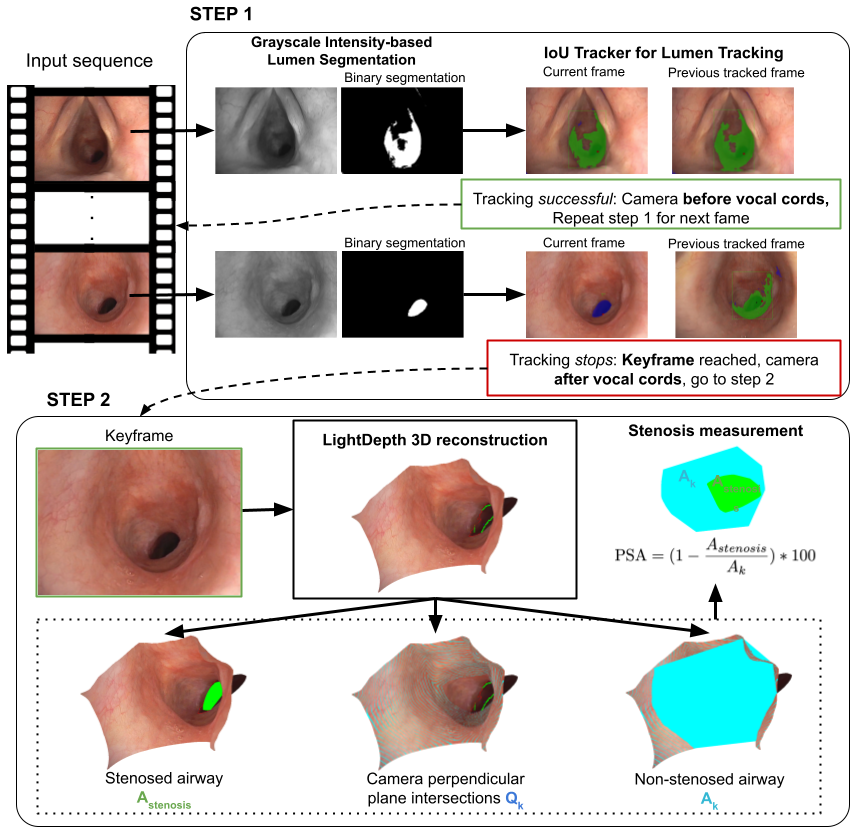}
    \caption{Overview of our proposed approach. The \textbf{first step} detects the darkest area through \textbf{grayscale intensity segmentation} and tracks it along the sequence using \textbf{IoU tracker}. Once the camera reaches the adequate location to measure the stenosis, the darkest segment changes shape significantly. The tracking is lost and the \textbf{keyframe} is reached. That keyframe is then passed to our \textbf{second step}, which produces a \textbf{3D reconstruction} of the airway using \textbf{LightDepth} model. The stenosed and non-stenosed airway areas are measured in that model.}
    \label{fig:metodo}
\end{figure}

Building on this foundation, our approach, presented in Fig.~\ref{fig:metodo}, first detects and tracks the airway lumen to select the optimal frame for stenosis evaluation. In that selected frame, we identify the stenosed region and estimate its severity by using a detailed 3D model generated based on the illumination decline. 

We use illumination decline for the two key steps of our proposed approach: 1) to track the deepest area in the image, corresponding to the airway lumen, to detect when the camera passes the vocal cords and to discern between stenosed and non-stenosed areas (see Section~\ref{met_seg}), and 2) to build a 3D model where we estimate the stenosis (see Section~\ref{met_rec}).

In the unique environment of endoscopic imaging, the co-location of camera and illumination source enables the recovery of geometric information through photometric analysis~\cite{Rodriguez-Puigvert_2023_ICCV,LightNeus,batlle2022photometric}. We observe that the color palette in bronchoscopic images is significantly influenced by the scene illumination. Therefore, we employ the spotlight light model of \cite{modrzejewski2020light}, later used in \cite{Rodriguez-Puigvert_2023_ICCV} to reconstruct the scene geometry by employing the illumination decline information. Moreover, in endoscopic imagery, the intensity of the image $\mathcal{I}$ decreases with the distance $d$, following the inverse square law and the gamma correction $\gamma$,
$\mathcal{I} \approx (1/d^2)^{1/\gamma}$. 
When the intensity falls below a certain intensity level, the available photometric information might no longer be sufficient for reliable 3D reconstruction. However, this information is useful for navigation during the procedure following the darkest area and identifying the stenosed tissue.

\subsection{Segmentation, Tracking and Frame Selection}\label{met_seg}
This step segments and tracks the darkest area along the sequence until it finds the best keyframe for subglottic stenosis estimation, just below the vocal cords. 
The darkest regions are segmented with a binary intensity threshold over the grayscale image.  
Then, we track the resulting segments with an IoU (Intersection Over Union) tracker~\cite{multiobjtracker_amd2018}. 
It is necessary to allow a few consecutive frames to fail in this tracking to make the approach robust to sudden illumination changes. 

When the tracker stops, it means that the darkest area segment has significantly changed, i.e., its IoU with the previous frame tracked segment is below the minimum IoU  established).  
At this point, we consider the camera has passed the vocal cords and reached the subglottis, and the darkest part of the image has shifted from corresponding to the vocal cords to the stenosis. This frame is selected as keyframe and used to run the second step of our pipeline. 

\subsection{3D Reconstruction and Stenosis Estimation}\label{met_rec}
Once the keyframe is selected, we compute the 3D reconstruction from this single view, employing the LightDepth model, as introduced by \cite{Rodriguez-Puigvert_2023_ICCV}. This model processes an input image ${I}$ and outputs an up-to-scale depth map $\widehat{d}$, an albedo map $\widehat{\rho}$, and normals ${\widehat{n}}$ derived from $\widehat{d}$.
LightDepth presents state-of-the-art results in other endoscopic settings, ideal for the illumination decline hypothesis, while experiments with DepthAnything~\cite{depthanything} and COLMAP~\cite{schoenberger2016mvs,schoenberger2016sfm} produced less detailed, noisier reconstructions.

Our 3D model is defined by a set of 3D points $X_{3D} = \pi^{-1}(\widehat{d},K)$, where $\pi^{-1}$ denotes the projection function, and $K$ represents the camera intrinsics.
As described in Section~\ref{met_seg}, we segment the stenosed area using illumination decline, i.e. we detect the darkest region in the image.
By segmenting the stenosed region in the image, and finding its correspondence in the obtained 3D model, we can precisely delineate the 3D contour of the stenotic region. This set of 3D points, $X_{3Dstenosis}$, outline the region and define a common plane, $\mathcal{P}$. We intersect $\mathcal{P}$ with the 3D model to obtain the area corresponding to the stenosis, $A_{stenosis}$ (see green region in the example from Fig.~\ref{fig:metodo} step 2):
\begin{equation}
    X_{3Dstenosis} = \{X \in X_{3D} | X \in \mathcal{P}\},
\end{equation}
In addition to delineating the stenosed area, we generate a series of planes $\mathcal{Q}_k$ oriented perpendicular to the camera to intersect with the 3D reconstruction $\mathcal{X_{3D}}$. The plane that results in the maximum intersection area $A_k$ is selected as the optimal reference area just beyond the vocal cords (blue region in the example from Fig.~\ref{fig:metodo} step 2). 
\begin{equation}
   k = \underset{k}{\text{argmax}} \, A_k
\end{equation}%
We quantify the degree of airway stenosis observed, $PSA$, by computing the ratio of the stenosed area  ($A_{stenosis}$) and the maximum area measured in the airway between the vocal cords and the stenosis ($A_k$). 
Additionally, to approximate the obstruction in terms of airway diameter, similar to the method proposed \cite{myer1994proposed}, we fit circles to the delineated stenosed and non stenosed areas in the 3D reconstruction, measure their diameters 
and compute the ratios ($PSD$).
\begin{align}
    \text{PSA} &= (1 - \frac{A_{stenosis}}{A_k})*100, & 
    \text{PSD} &= (1-\frac{{diameter}_{stenosis}}{{diameter}_k})*100
\end{align}

\subsection{Subglottic Stenosis Dataset}\label{met_data}
Our Subglottic Stenosis (SGS) Dataset encompasses 16 bronchoscopy videos from 11 patients (A through J), recorded during real medical procedures using three bronchoscopes: Olympus BF-H1100, Olympus BRF-180 and Olympus BF-1TH1100. Table~\ref{tab:dataset} summarizes the dataset contents, with examples in Fig.~\ref{fig:ex_stenosis}.
For Patients A, C and D, we include CT scans with ground-truth stenosis measurements in terms of areas, obtained from intersections between a predefined plane — at the trachea's widest point and stenosis location —and the 3D CT reconstruction. For patients A, B and D to J, obstruction percentage visually estimated by experts - by comparing estimated stenosis diameter to average airway diameter without stenosis as done in~\cite{myer1994proposed} - is available as reference. Patients G, H and I are healthy without stenosis. Patients B, D and E have multiple bronchoscopy videos under identical pathological conditions by different physicians.

\begin{table}[tb]
    \setlength{\tabcolsep}{6pt}
    
    \centering
    \caption{Overview of the released dataset. 16 bronchoscopy videos from 11 patients. 12 sequences with reference obstruction percentage estimation from experts (\textbf{Expert estimation}), 5 with corresponding CT scan. Repeated procedures for patients B, D and E from different physicians.}
    \begin{tabular}{c c ccc c ccc cc c c c c c c c}
    \toprule
       \textbf{Sequence}  & 1 & 2 & 3 & 4 & 5 & 6 & 7 & 8 & 9 & 10 & 11 & 12 & 13 & 14 & 15 & 16 \\ \midrule
       \textbf{Patient ID} & A & B & B & B & C & D & D & D & E & E & F & G & H & I & J & K \\ \midrule
       \textbf{CT scan} & \checkmark & \ding{55} & \ding{55} & \ding{55} & \checkmark & \checkmark & \checkmark & \checkmark & \ding{55} & \ding{55} & \ding{55} & \ding{55} & \ding{55} & \ding{55} & \ding{55} & \ding{55} \\\midrule       
       \textbf{Expert} & \multirow{2}{*}{\checkmark} & \multirow{2}{*}{\checkmark}  & \multirow{2}{*}{\checkmark}  & \multirow{2}{*}{\ding{55}}  & \multirow{2}{*}{\ding{55}}  & \multirow{2}{*}{\checkmark}  & \multirow{2}{*}{\checkmark}  &  \multirow{2}{*}{\ding{55}} &  \multirow{2}{*}{\checkmark} &  \multirow{2}{*}{\ding{55}} & \multirow{2}{*}{\checkmark} & \multirow{2}{*}{\checkmark} & \multirow{2}{*}{\checkmark} & \multirow{2}{*}{\checkmark} & \multirow{2}{*}{\checkmark} & \multirow{2}{*}{\checkmark}  \\ 
       \textbf{estimation} &  &  &  &  &  &  &  &  &  &  & & & & &  &\\ \midrule
       \textbf{Scope ID} & 1 & 1 & 1 & 1 & 1 & 2 & 2 & 2 & 2 & 2 & 3 & 3 & 1 & 3 & 1 & 3 \\
    \bottomrule
    \end{tabular}
    \label{tab:dataset}
\end{table}
\begin{figure}[tb]
    
    \centering
    \setlength{\tabcolsep}{0.5pt}
    \begin{tabular}{cccccc}
     A & B & C & D & G & H \\
     \includegraphics[width=0.15\linewidth]{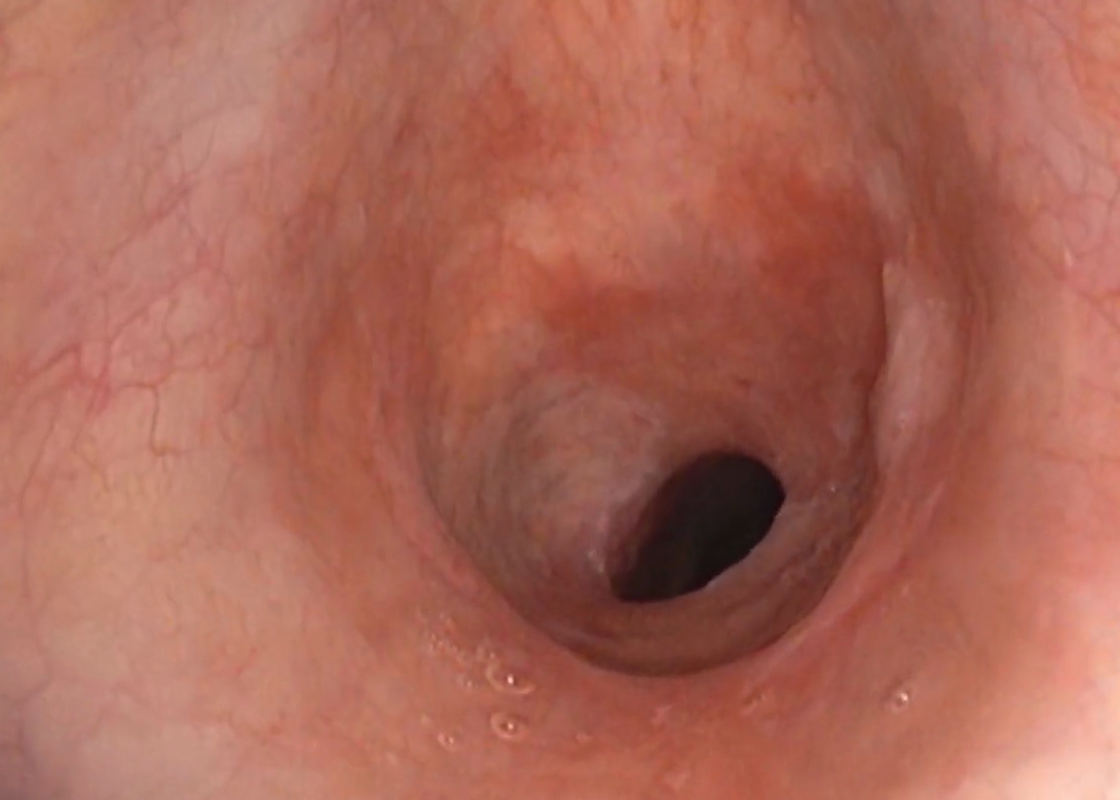} &
     \includegraphics[width=0.15\linewidth]{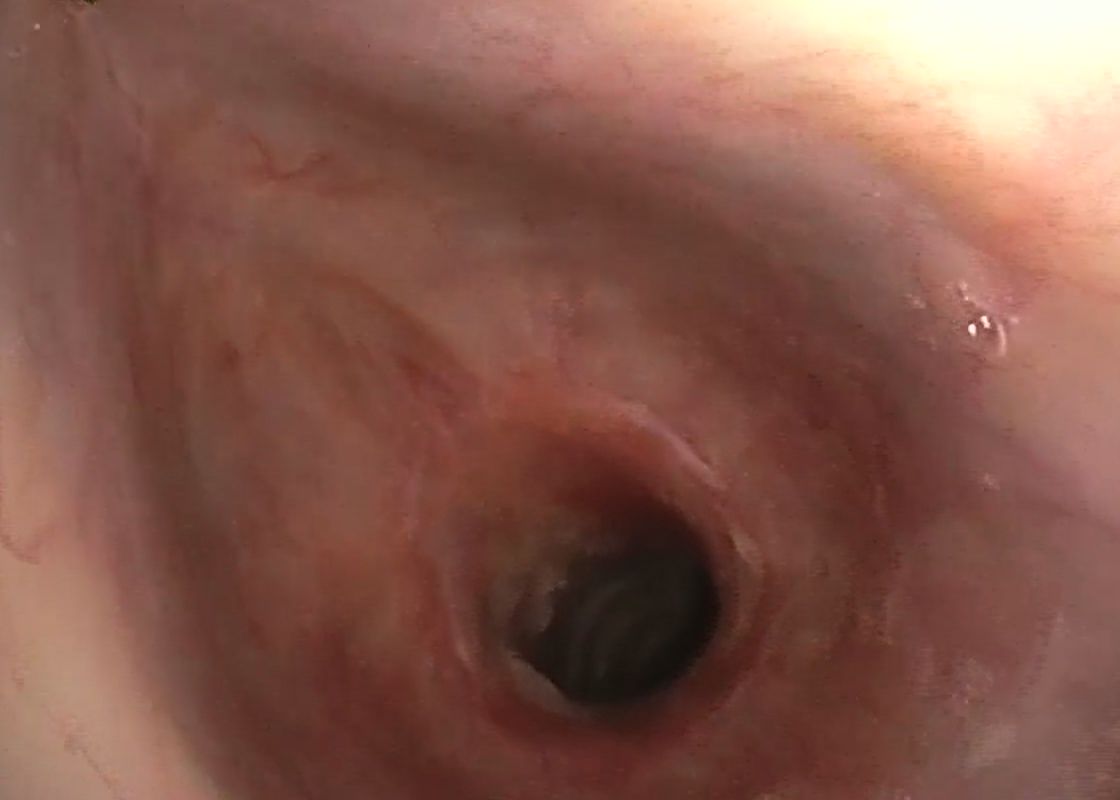} &
      \includegraphics[width=0.15\linewidth]{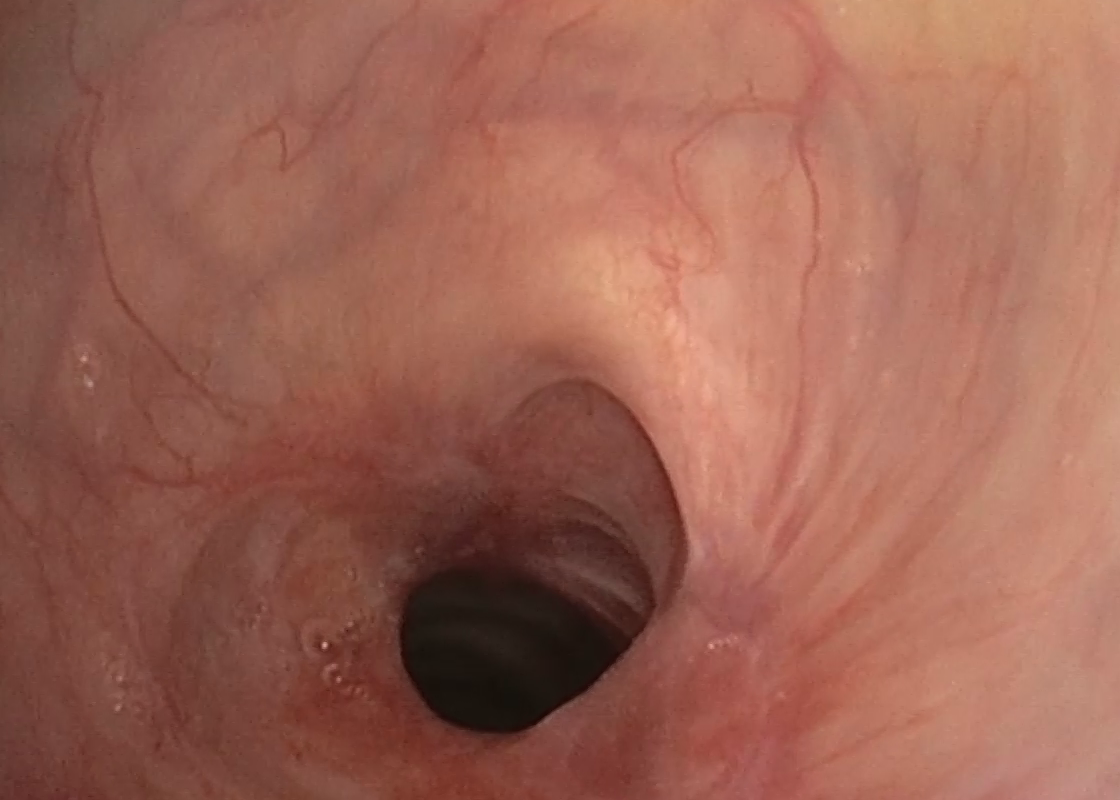} &
     \includegraphics[width=0.15\linewidth]{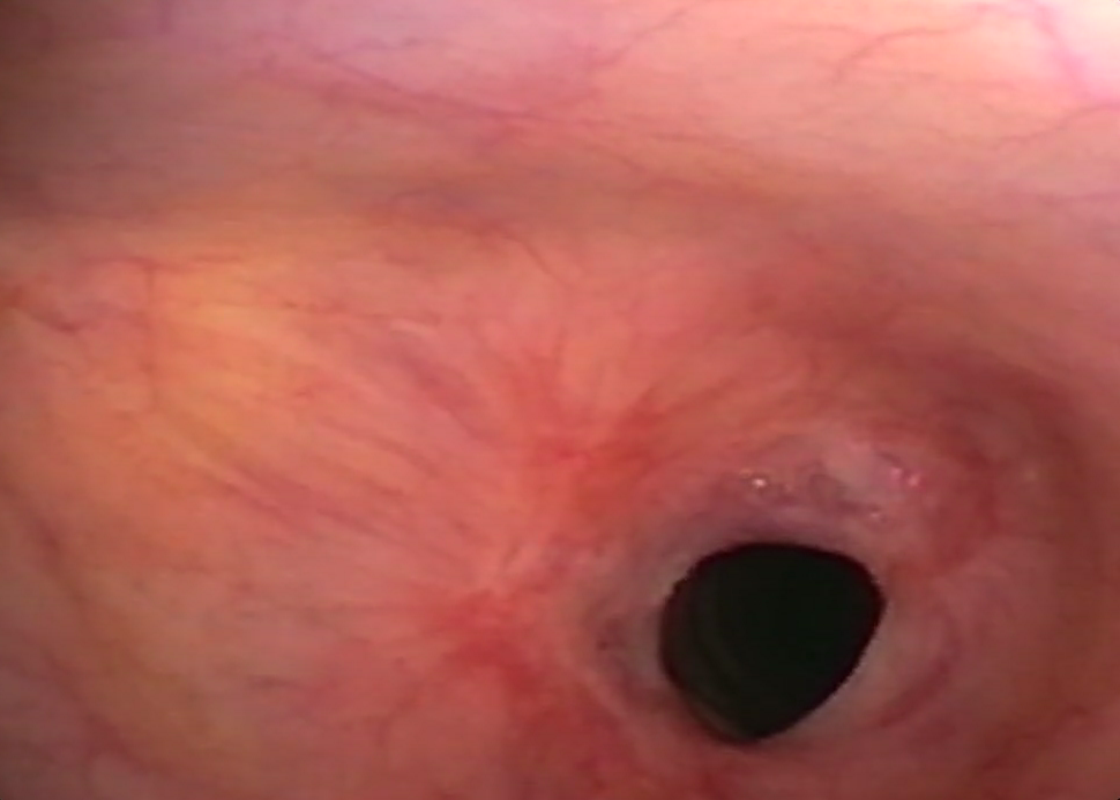}& 
     \includegraphics[width=0.15\linewidth]{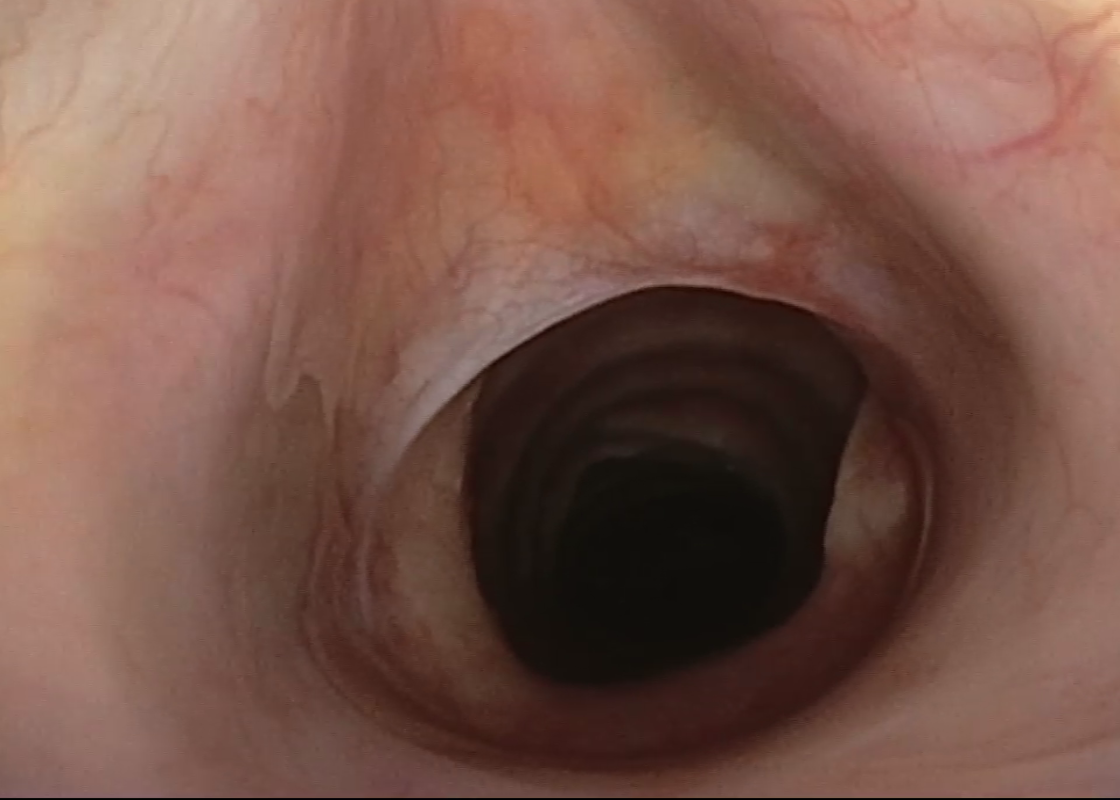}&
     \includegraphics[width=0.15\linewidth]{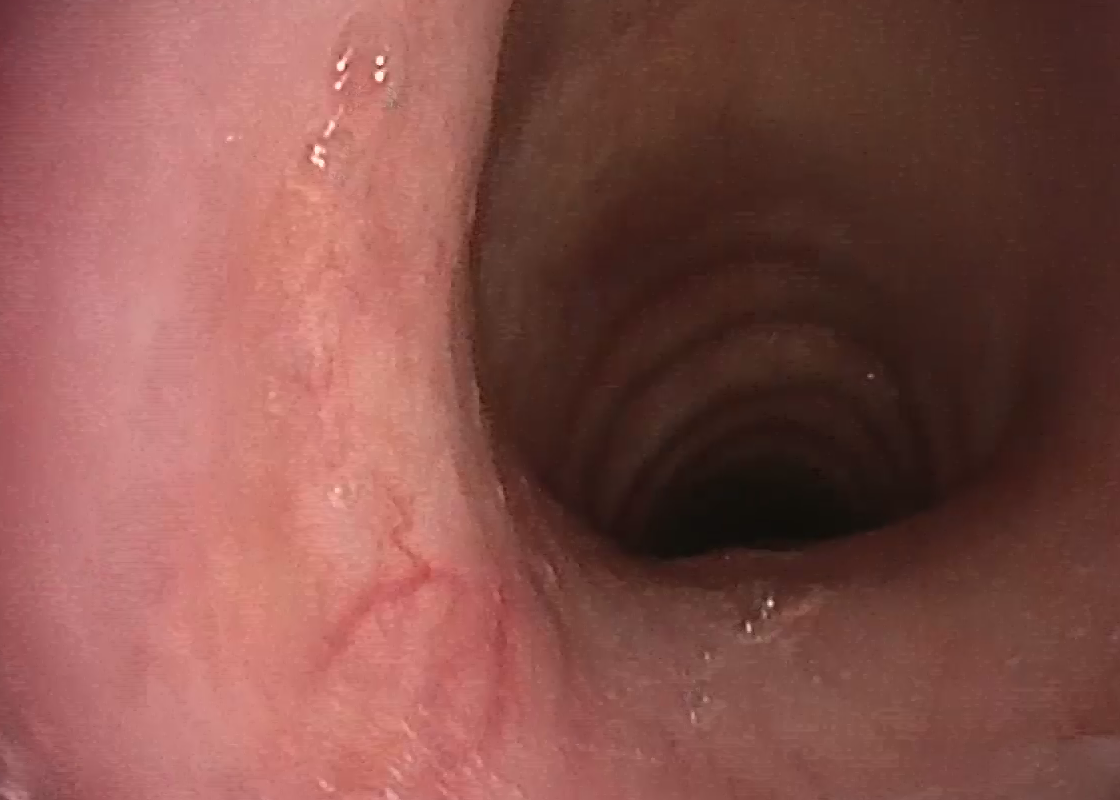}\\
    \end{tabular}
    \caption{Examples of subglottic area appearance for patients A, B, C, D, G and H from our proposed dataset. Patients A, B, C and D present varying degrees of stenosis. Patients G and H are healthy, and the trachea with its characteristic cartilaginous rings can be seen in the back.}
    \label{fig:ex_stenosis}
\end{figure}

\section{Experiments}

\subsection{Experimental Setup}

\noindent\textbf{Segmentation and Tracking parameters.} Segmentation involves detecting the pixels in the grayscale image with intensity below 50, chosen following the illumination decline and the inverse square law, described in Section~\ref{sec:overview}, and tested though visual assessment on real colonoscopy recordings from EndoMapper dataset~\cite{azagra2023endomapper}. This threshold remains effective despite small illumination variations within each bronchoscopy sequence.
Tracking relies on a minimum IoU of 50\% for detection and 25 frames before a tracked segment is lost.

\noindent\textbf{3D Reconstruction model training.}
The 3D reconstruction at the keyframe is obtained using LightDepth~\cite{Rodriguez-Puigvert_2023_ICCV}, in its LightDepth~\cite{Rodriguez-Puigvert_2023_ICCV} U-Net~\cite{Ronneberger2015} version for faster inference time. 
We train LightDepth on 2800 frames from sequence 1 of the Subglottic Stenosis dataset, covering various anatomical regions, including the subglottic and tracheal areas, with the corresponding geometric and photometric calibration of the bronchoscope. This sequence is used for LightDepth training, but is kept to evaluate the quality of stenosis estimation, as it includes scarce CT ground-truth data and that errors in this sequence are similar than in the rest.
LightDepth is trained for 5 epochs using a learning rate of $lr=10^{-4}$, a depth smoothness regularization weight of $\lambda_s=0.1$, and a specularity-aware loss component with $\lambda_{sp}=1$.
The encoder is initialized from ImageNet pre-trained weights. The albedo and depth decoders are trained from scratch. 

\noindent\textbf{Metrics.}
The main metrics computed to evaluate the accuracy and consistency of our approach are: 
\begin{itemize}
    \item Percentage of correctly selected keyframes (\textbf{Correct Keyframes}), which measures if the keyframe selected, using our first step (Section~\ref{met_seg}), falls within a manually defined optimal frame interval, spanning from when the bronchoscope is just below the vocal cords to just before reaching the stenosis.
    \item Value range and Absolute Difference (\textbf{Diff.}) between 
     estimations (PSD or PSA) computed on sequences of the same patient and condition.
    \item Absolute Error (AE) for each sequence and Mean Absolute Error (\textbf{MAE}) to quantify the average error in estimated PSA and PSD for all sequences with PSA or PSD reference. 
Given a sequence $i$ and its corresponding ground-truth $GT_i$ and computed estimation $ce_i$, the MAE over N sequences is computed as follows:

\begin{equation}
    MAE = \frac{\sum_{i=1}^N|GT_i - ce_i|}{N} = \frac{\sum_{i=1}^NAE_{i}}{N}.
\end{equation}
\end{itemize}

\subsection{Results}
This section presents the quantitative evaluation, regarding accuracy and consistency, of our full proposed pipeline. The supplementary video in Online Resource 2 shows how our pipeline runs on sequences 1 and 2 of the SGS dataset.\\

\noindent{\textbf{Comparison to CT scan and expert estimations.}} First, we compare the stenosis estimations obtained using our pipeline, on the 16 sequences of SGS dataset, to the CT scan PSA ground-truth and/or expert PSD estimations.

As no baseline is available for the task of SGS estimation, up to our knowledge, we implement several baselines to better illustrate the performance of our approach. In particular, we consider several alternatives for the critical initial stage of keyframe selection. We replace our intensity-based segmentation (\textit{\textbf{Ours}}, Section~\ref{met_seg}) with well known segmentation alternatives to achieve the target (lumen) segment, maintaining the rest of the keyframe selection process (Section~\ref{met_seg}, tracking and end-point selection) and 3D reconstruction and estimation step (Section~\ref{met_rec}) as State-of-the-Art methods. The alternative baselines considered are :
\begin{itemize}
\item{\textbf{SAM-based}}, uses SAM~\cite{kirillov2023segment} segmentation prompted with the darkest pixel. 
\item{\textbf{SLIC-based}}, uses SLIC~\cite{achanta2010slic} superpixel segmentation, selecting the segment that contains the darkest pixel.
\item{\textbf{Manual}}, manually selecting the first frame right after the bronchoscope passes the vocal cords, as a reference lower bound for the error of our second stage (Section~\ref{met_rec}).
\end{itemize}

Table~\ref{tab:gt_meas} presents the average results obtained over all sequences. Detailed results per sequence are shown in Tables 1 and 2 of Online Resource 1. 
Our pipeline, using the proposed intensity-based segmentation, consistently selects keyframes within the ground-truth interval for all sequences, while the other alternatives present several failure cases. 
Moreover, our proposed approach has significantly lower PSA and PSD error, and achieves the closest error to that obtained when manually given the best keyframe possible. 
Additionally, paired t-tests between ground-truth PSA (reference expert PSD respectively) and corresponding estimates obtained using our full method, with p-value 0.6039 (0.2766 respectively), greater than 0.05, shows no statistically significant difference between means. Overall, this experiment highlights the ability of our approach to accurately estimate the stenosis. 

\begin{table}[tb]
    
    \setlength{\tabcolsep}{3pt}
    \centering
    \caption{Proposed stenosis estimation (Ours)  compared to other baselines with different keyframe selection approaches. 
    Mean Absolute Error (MAE) is computed for \textbf{\textit{all tests}} but also considering only tests where the selected keyframes are correct (\textbf{Correct KF only}). 
    }
    \begin{tabular}{cccccc}
    \toprule
         \textbf{Keyframe } 
         & \textbf{Correct } & \multicolumn{2}{c}{\textbf{PSA MAE (\%)}} &  \multicolumn{2}{c}{\textbf{PSD MAE (\%)}} \\
         \textbf{Selection}
         & \textbf{Keyframes (\%)} & \textbf{Correct KF only} & \textbf{ All tests} & \textbf{Correct KF only} & \textbf{ All tests} \\
         
       \toprule
       SAM-based
       & 37,5 & 6,43 & 39,84 & 15,33 & 34,72\\
       \midrule
       SLIC-based 
       & 25 & 16,07 & 52,40 & 5,34 & 28,44\\
       \midrule
       Ours
       & \textbf{100} & 4,84 & \textbf{4,84} & 6,70 & \textbf{6,70} \\
       \midrule
       \midrule
       Manual
       & 100 & 1,21 & 1,21 & 5,31 & 5,31 \\
    \bottomrule 
    \end{tabular}
    \label{tab:gt_meas}
\end{table}

\noindent\textbf{Same patient and interexpert consistency.}
This second experiment analyzes the consistency of our estimations in the SGS Dataset. This analysis can only use sequences corresponding to patients (B, D and E) with several procedures without patient condition changes between the sequences. The PSA and PSD for both sequences of the same patient should be similar or close in range. For patients B and D, two expert PSD estimations are also available, and their variations are shown as reference. 
Table~\ref{tab:gt_var} presents estimation range and difference between minimum and maximum estimations.  
Using our approach, estimations differ on average 1,79\% in PSA and 5,07\% in PSD between the two sequences, while expert PSD estimates vary on average 7,5\%. 
Overall, our pipeline shows consistency over repeated estimations for the same patient on comparable ranges than expert estimations. 

\begin{table}[tb]
    
    \setlength{\tabcolsep}{14pt}
    \centering
    \caption{Stenosis estimation consistency analysis. PSA and PSD estimation range (\textbf{Range}) and difference (\textbf{Diff.}) between estimations on two sequences corresponding to the same patient (\textbf{S.-P.}). Estimates obtained using two methods: our approach for PSA and PSD, and from expert estimations for PSD (patients B and D).}
    \begin{tabular}{cccc}
    \toprule
         \textbf{S.-P.} & \textbf{Method} & \textbf{PSA Range (Diff.)  (\%)} & \textbf{PSD Range (Diff.)  (\%)}  \\
       \toprule
       2,3-B & Ours & 58,51-62,16 (3,65) & 62,94-55,70 (7,24)\\
       2,3-B & Experts & / & 55-60 (5) \\
       \midrule
       6,7-D & Ours & 61,90-62,35 (0,45) & 57,89-59,09 (1,20) \\
       6,7-D & Experts & / & 60-70 (10)\\
       \midrule
       9,10-E & Ours & 64,58-65,85 (1,27) & 58,95-65,71 (6,76) \\
    \bottomrule 
    \end{tabular}
    \label{tab:gt_var}
\end{table}

\section{Conclusion and Discussion}
In this work, we have presented a pipeline to automatically evaluate subglottic stenosis using only bronchoscopy videos, without traversing the stenosed region. Our two step approach first selects the optimal frame for measuring the stenosis through segmentation and tracking, and then uses a dynamically computed 3D airway reconstruction to measure the stenosis obstruction at that keyframe.

We evaluate our approach using our novel benchmark for subglottic stenosis estimation, the first of its kind being publicly released to facilitate reproducibility and encourages further research.
Our results show that our pipeline provides an assistance tool in real procedures for first robust stenosis severity estimation and to monitor the condition during follow up procedures, to assess the stenosis evolution. Our method standardizes stenosis measurement, reducing the subjectivity traditionally observed, and easily integrates into existing bronchoscopy workflows without needing extra hardware or doctor training. Moreover, measuring at the keyframe takes 7s on average on a GPU GeForce RTX3090, and could run in parallel to doctor assessment.

\begin{figure}[tb]
    \centering
    \includegraphics[width=0.55\linewidth]{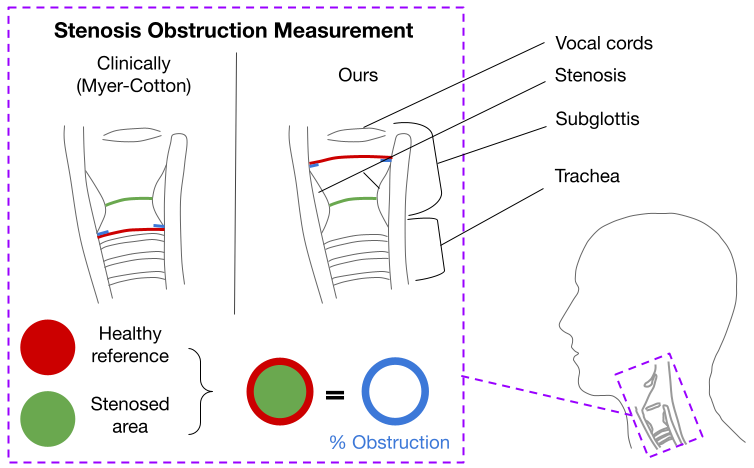}
    \caption{Obstruction measurement as done clinically (Myer-Cotton classification) compared to our approach. The green area represents the stenosed area, for which the measurement is the same clinically and in our case. The red area corresponds to the healthy reference airway: clinically measured in the trachea below the stenosis, in our approach measured between the vocal cords and the stenosis. The obstruction (in blue) is the difference between the healthy and stenosed areas.}
    \label{fig:refMC}
\end{figure}

However, there are few limitations to address in future work. 
Our method slightly underestimates the stenosis index (SI) and our healthy reference measurement location differs from the clinically used Myer-Cotton (MC) classification~\cite{myer1994proposed}, as seen in Fig.~\ref{fig:refMC}. We measure the largest airway area right after the vocal cords, before traversing the stenosis, while the MC classification typically considers the area of the trachea after the stenosis. 
Future work thus includes collecting and releasing more expert-supervised data to calibrate our approach and validate a direct comparison with the MC classification and possibly correct the slight underestimation. 

Our work paves the way for future research using 3D geometric information to reduce subjectivity and procedure time, benefiting patients and physicians in diagnosis and monitoring.

\section*{Statements and Declarations}
\textbf{Conflict of interest.} The authors declare that they have no conflict of interest.
\noindent\textbf{Funding.} This project has received funding from the European Union’s Horizon 2020 research and innovation programme under grant agreement No 863146, from the Aragon Government project FSE-T45 23R, and from AEI and the European Union under grant agreement PID2021-125514NB-I00.

\noindent\textbf{Acknowledgments.} 
The authors express their gratitude to the medical staff of Hospital Universitario Miguel Servet for their support in obtaining bronchoscopy procedure data, with special thanks to Elisa Minchole.
\bibliography{bibliography}


\begin{thebibliography}{24}
\ifx \bisbn   \undefined \def \bisbn  #1{ISBN #1}\fi
\ifx \binits  \undefined \def \binits#1{#1}\fi
\ifx \bauthor  \undefined \def \bauthor#1{#1}\fi
\ifx \batitle  \undefined \def \batitle#1{#1}\fi
\ifx \bjtitle  \undefined \def \bjtitle#1{#1}\fi
\ifx \bvolume  \undefined \def \bvolume#1{\textbf{#1}}\fi
\ifx \byear  \undefined \def \byear#1{#1}\fi
\ifx \bissue  \undefined \def \bissue#1{#1}\fi
\ifx \bfpage  \undefined \def \bfpage#1{#1}\fi
\ifx \blpage  \undefined \def \blpage #1{#1}\fi
\ifx \burl  \undefined \def \burl#1{\textsf{#1}}\fi
\ifx \doiurl  \undefined \def \doiurl#1{\url{https://doi.org/#1}}\fi
\ifx \betal  \undefined \def \betal{\textit{et al.}}\fi
\ifx \binstitute  \undefined \def \binstitute#1{#1}\fi
\ifx \binstitutionaled  \undefined \def \binstitutionaled#1{#1}\fi
\ifx \bctitle  \undefined \def \bctitle#1{#1}\fi
\ifx \beditor  \undefined \def \beditor#1{#1}\fi
\ifx \bpublisher  \undefined \def \bpublisher#1{#1}\fi
\ifx \bbtitle  \undefined \def \bbtitle#1{#1}\fi
\ifx \bedition  \undefined \def \bedition#1{#1}\fi
\ifx \bseriesno  \undefined \def \bseriesno#1{#1}\fi
\ifx \blocation  \undefined \def \blocation#1{#1}\fi
\ifx \bsertitle  \undefined \def \bsertitle#1{#1}\fi
\ifx \bsnm \undefined \def \bsnm#1{#1}\fi
\ifx \bsuffix \undefined \def \bsuffix#1{#1}\fi
\ifx \bparticle \undefined \def \bparticle#1{#1}\fi
\ifx \barticle \undefined \def \barticle#1{#1}\fi
\bibcommenthead
\ifx \bconfdate \undefined \def \bconfdate #1{#1}\fi
\ifx \botherref \undefined \def \botherref #1{#1}\fi
\ifx \url \undefined \def \url#1{\textsf{#1}}\fi
\ifx \bchapter \undefined \def \bchapter#1{#1}\fi
\ifx \bbook \undefined \def \bbook#1{#1}\fi
\ifx \bcomment \undefined \def \bcomment#1{#1}\fi
\ifx \oauthor \undefined \def \oauthor#1{#1}\fi
\ifx \citeauthoryear \undefined \def \citeauthoryear#1{#1}\fi
\ifx \endbibitem  \undefined \def \endbibitem {}\fi
\ifx \bconflocation  \undefined \def \bconflocation#1{#1}\fi
\ifx \arxivurl  \undefined \def \arxivurl#1{\textsf{#1}}\fi
\csname PreBibitemsHook\endcsname

\bibitem[\protect\citeauthoryear{Begnaud et~al.}{2015}]{begnaud2015measuring}
\begin{barticle}
\bauthor{\bsnm{Begnaud}, \binits{A.}},
\bauthor{\bsnm{Connett}, \binits{J.E.}},
\bauthor{\bsnm{Harwood}, \binits{E.M.}},
\bauthor{\bsnm{Jantz}, \binits{M.A.}},
\bauthor{\bsnm{Mehta}, \binits{H.J.}}:
\batitle{Measuring central airway obstruction. what do bronchoscopists do?}
\bjtitle{Annals of the American Thoracic Society}
\bvolume{12}(\bissue{1}),
\bfpage{85}--\blpage{90}
(\byear{2015})
\end{barticle}
\endbibitem

\bibitem[\protect\citeauthoryear{Murgu and Colt}{2013}]{murgu2013subjective}
\begin{barticle}
\bauthor{\bsnm{Murgu}, \binits{S.}},
\bauthor{\bsnm{Colt}, \binits{H.}}:
\batitle{Subjective assessment using still bronchoscopic images misclassifies airway narrowing in laryngotracheal stenosis}.
\bjtitle{Interactive cardiovascular and thoracic surgery}
\bvolume{16}(\bissue{5}),
\bfpage{655}--\blpage{660}
(\byear{2013})
\end{barticle}
\endbibitem

\bibitem[\protect\citeauthoryear{Rodr{\'\i}guez-Puigvert et~al.}{2023}]{Rodriguez-Puigvert_2023_ICCV}
\begin{bchapter}
\bauthor{\bsnm{Rodr{\'\i}guez-Puigvert}, \binits{J.}},
\bauthor{\bsnm{Batlle}, \binits{V.M.}},
\bauthor{\bsnm{Montiel}, \binits{J.M.M.}},
\bauthor{\bsnm{Martinez-Cantin}, \binits{R.}},
\bauthor{\bsnm{Fua}, \binits{P.}},
\bauthor{\bsnm{Tard\'os}, \binits{J.D.}},
\bauthor{\bsnm{Civera}, \binits{J.}}:
\bctitle{Lightdepth: Single-view depth self-supervision from illumination decline}.
In: \bbtitle{IEEE/CVF Int. Conf. on Computer Vision (ICCV)},
pp. \bfpage{21273}--\blpage{21283}
(\byear{2023})
\end{bchapter}
\endbibitem

\bibitem[\protect\citeauthoryear{Sharma et~al.}{2016}]{sharma2016quantitative}
\begin{barticle}
\bauthor{\bsnm{Sharma}, \binits{G.K.}},
\bauthor{\bsnm{Chin~Loy}, \binits{A.}},
\bauthor{\bsnm{Su}, \binits{E.}},
\bauthor{\bsnm{Jing}, \binits{J.}},
\bauthor{\bsnm{Chen}, \binits{Z.}},
\bauthor{\bsnm{Wong}, \binits{B.J.}},
\bauthor{\bsnm{Verma}, \binits{S.}}:
\batitle{Quantitative evaluation of adult subglottic stenosis using intraoperative long-range optical coherence tomography}.
\bjtitle{Annals of Otology, Rhinology \& Laryngology}
\bvolume{125}(\bissue{10}),
\bfpage{815}--\blpage{822}
(\byear{2016})
\end{barticle}
\endbibitem

\bibitem[\protect\citeauthoryear{Banach et~al.}{2023}]{banach2023computer}
\begin{barticle}
\bauthor{\bsnm{Banach}, \binits{A.}},
\bauthor{\bsnm{Naito}, \binits{M.}},
\bauthor{\bsnm{King}, \binits{F.}},
\bauthor{\bsnm{Masaki}, \binits{F.}},
\bauthor{\bsnm{Tsukada}, \binits{H.}},
\bauthor{\bsnm{Hata}, \binits{N.}}:
\batitle{Computer-based airway stenosis quantification from bronchoscopic images: preliminary results from a feasibility trial}.
\bjtitle{International Journal of Computer Assisted Radiology and Surgery}
\bvolume{18}(\bissue{4}),
\bfpage{707}--\blpage{713}
(\byear{2023})
\end{barticle}
\endbibitem

\bibitem[\protect\citeauthoryear{S{\'a}nchez et~al.}{2015}]{sanchez2015toward}
\begin{barticle}
\bauthor{\bsnm{S{\'a}nchez}, \binits{C.}},
\bauthor{\bsnm{Bernal}, \binits{J.}},
\bauthor{\bsnm{S{\'a}nchez}, \binits{F.J.}},
\bauthor{\bsnm{Diez}, \binits{M.}},
\bauthor{\bsnm{Rosell}, \binits{A.}},
\bauthor{\bsnm{Gil}, \binits{D.}}:
\batitle{Toward online quantification of tracheal stenosis from videobronchoscopy}.
\bjtitle{International journal of computer assisted radiology and surgery}
\bvolume{10},
\bfpage{935}--\blpage{945}
(\byear{2015})
\end{barticle}
\endbibitem

\bibitem[\protect\citeauthoryear{Keuth et~al.}{2023}]{keuth2023weakly}
\begin{bchapter}
\bauthor{\bsnm{Keuth}, \binits{R.}},
\bauthor{\bsnm{Heinrich}, \binits{M.}},
\bauthor{\bsnm{Eichenlaub}, \binits{M.}},
\bauthor{\bsnm{Himstedt}, \binits{M.}}:
\bctitle{Weakly supervised airway orifice segmentation in video bronchoscopy}.
In: \bbtitle{Medical Imaging 2023: Image Processing},
vol. \bseriesno{12464},
pp. \bfpage{58}--\blpage{65}
(\byear{2023}).
\bcomment{SPIE}
\end{bchapter}
\endbibitem

\bibitem[\protect\citeauthoryear{S{\'a}nchez et~al.}{2014}]{sanchez2014line}
\begin{bchapter}
\bauthor{\bsnm{S{\'a}nchez}, \binits{C.}},
\bauthor{\bsnm{Bernal}, \binits{J.}},
\bauthor{\bsnm{Gil}, \binits{D.}},
\bauthor{\bsnm{S{\'a}nchez}, \binits{F.J.}}:
\bctitle{On-line lumen centre detection in gastrointestinal and respiratory endoscopy}.
In: \bbtitle{CLIP 2013, Held in Conjunction with MICCAI 2013},
pp. \bfpage{31}--\blpage{38}
(\byear{2014}).
\bcomment{Springer}
\end{bchapter}
\endbibitem

\bibitem[\protect\citeauthoryear{Achanta et~al.}{June 2010}]{achanta2010slic}
\begin{botherref}
\oauthor{\bsnm{Achanta}, \binits{R.}},
\oauthor{\bsnm{Shaji}, \binits{A.}},
\oauthor{\bsnm{Smith}, \binits{K.}},
\oauthor{\bsnm{Lucchi}, \binits{A.}},
\oauthor{\bsnm{Fua}, \binits{P.}},
\oauthor{\bsnm{S{\"u}sstrunk}, \binits{S.}}:
Slic superpixels.
EPFL Technical Report 149300
(June 2010)
\end{botherref}
\endbibitem

\bibitem[\protect\citeauthoryear{Kirillov et~al.}{2023}]{kirillov2023segment}
\begin{botherref}
\oauthor{\bsnm{Kirillov}, \binits{A.}},
\oauthor{\bsnm{Mintun}, \binits{E.}},
\oauthor{\bsnm{Ravi}, \binits{N.}},
\oauthor{\bsnm{Mao}, \binits{H.}},
\oauthor{\bsnm{Rolland}, \binits{C.}},
\oauthor{\bsnm{Gustafson}, \binits{L.}},
\oauthor{\bsnm{Xiao}, \binits{T.}},
\oauthor{\bsnm{Whitehead}, \binits{S.}},
\oauthor{\bsnm{Berg}, \binits{A.C.}},
\oauthor{\bsnm{Lo}, \binits{W.-Y.}},
\oauthor{\bsnm{Doll{\'a}r}, \binits{P.}},
\oauthor{\bsnm{Girshick}, \binits{R.}}:
Segment anything.
IEEE/CVF Int. Conf. on Computer Vision,
4015--4026
(2023)
\end{botherref}
\endbibitem

\bibitem[\protect\citeauthoryear{Murgu and Colt}{2009}]{murgu2009morphometric}
\begin{barticle}
\bauthor{\bsnm{Murgu}, \binits{S.}},
\bauthor{\bsnm{Colt}, \binits{H.G.}}:
\batitle{Morphometric bronchoscopy in adults with central airway obstruction: case illustrations and review of the literature}.
\bjtitle{The Laryngoscope}
\bvolume{119}(\bissue{7}),
\bfpage{1318}--\blpage{1324}
(\byear{2009})
\end{barticle}
\endbibitem

\bibitem[\protect\citeauthoryear{Francom et~al.}{2019}]{francom2019clinical}
\begin{barticle}
\bauthor{\bsnm{Francom}, \binits{C.R.}},
\bauthor{\bsnm{Best}, \binits{C.A.}},
\bauthor{\bsnm{Eaton}, \binits{R.G.}},
\bauthor{\bsnm{Pepper}, \binits{V.}},
\bauthor{\bsnm{Onwuka}, \binits{A.J.}},
\bauthor{\bsnm{Breuer}, \binits{C.K.}},
\bauthor{\bsnm{Lind}, \binits{M.N.M.}},
\bauthor{\bsnm{Grischkan}, \binits{J.M.}},
\bauthor{\bsnm{Chiang}, \binits{T.}}:
\batitle{Clinical validation and reproducibility of endoscopic airway measurement in pediatric aerodigestive evaluation}.
\bjtitle{International journal of pediatric otorhinolaryngology}
\bvolume{116},
\bfpage{65}--\blpage{69}
(\byear{2019})
\end{barticle}
\endbibitem

\bibitem[\protect\citeauthoryear{Visentini-Scarzanella et~al.}{2017}]{visentini2017deep}
\begin{barticle}
\bauthor{\bsnm{Visentini-Scarzanella}, \binits{M.}},
\bauthor{\bsnm{Sugiura}, \binits{T.}},
\bauthor{\bsnm{Kaneko}, \binits{T.}},
\bauthor{\bsnm{Koto}, \binits{S.}}:
\batitle{Deep monocular 3d reconstruction for assisted navigation in bronchoscopy}.
\bjtitle{International journal of computer assisted radiology and surgery}
\bvolume{12},
\bfpage{1089}--\blpage{1099}
(\byear{2017})
\end{barticle}
\endbibitem

\bibitem[\protect\citeauthoryear{Azagra et~al.}{2023}]{azagra2023endomapper}
\begin{barticle}
\bauthor{\bsnm{Azagra}, \binits{P.}},
\bauthor{\bsnm{Sostres}, \binits{C.}},
\bauthor{\bsnm{Ferr{\'a}ndez}, \binits{{\'A}.}},
\bauthor{\bsnm{Riazuelo}, \binits{L.}},
\bauthor{\bsnm{Tomasini}, \binits{C.}},
\bauthor{\bsnm{Barbed}, \binits{O.L.}},
\bauthor{\bsnm{Morlana}, \binits{J.}},
\bauthor{\bsnm{Recasens}, \binits{D.}},
\bauthor{\bsnm{Batlle}, \binits{V.M.}},
\bauthor{\bsnm{G{\'o}mez-Rodr{\'\i}guez}, \binits{J.J.}},
\bauthor{\bsnm{Elvira}, \binits{R.}},
\bauthor{\bsnm{L{\'o}pez}, \binits{J.}},
\bauthor{\bsnm{Oriol}, \binits{C.}},
\bauthor{\bsnm{Civera}, \binits{J.}},
\bauthor{\bsnm{Tard{\'o}s}, \binits{J.}},
\bauthor{\bsnm{Murillo}, \binits{A.C.}},
\bauthor{\bsnm{Lanas}, \binits{{\'A}.}},
\bauthor{\bsnm{Montiel}, \binits{J.M.M.}}:
\batitle{Endomapper dataset of complete calibrated endoscopy procedures}.
\bjtitle{Scientific Data}
\bvolume{10}(\bissue{1}),
\bfpage{671}
(\byear{2023})
\end{barticle}
\endbibitem

\bibitem[\protect\citeauthoryear{Pogorelov et~al.}{2017}]{pogorelov2017kvasir}
\begin{bchapter}
\bauthor{\bsnm{Pogorelov}, \binits{K.}},
\bauthor{\bsnm{Randel}, \binits{K.R.}},
\bauthor{\bsnm{Griwodz}, \binits{C.}},
\bauthor{\bsnm{Eskeland}, \binits{S.L.}},
\bauthor{\bsnm{Lange}, \binits{T.}},
\bauthor{\bsnm{Johansen}, \binits{D.}},
\bauthor{\bsnm{Spampinato}, \binits{C.}},
\bauthor{\bsnm{Dang-Nguyen}, \binits{D.-T.}},
\bauthor{\bsnm{Lux}, \binits{M.}},
\bauthor{\bsnm{Schmidt}, \binits{P.T.}},
\bauthor{\bsnm{Riegler}, \binits{M.}},
\bauthor{\bsnm{Halvorsen}, \binits{P.}}:
\bctitle{Kvasir: A multi-class image dataset for computer aided gastrointestinal disease detection}.
In: \bbtitle{8th ACM on Multimedia Systems Conference},
pp. \bfpage{164}--\blpage{169}
(\byear{2017})
\end{bchapter}
\endbibitem

\bibitem[\protect\citeauthoryear{Batlle et~al.}{2023}]{LightNeus}
\begin{bchapter}
\bauthor{\bsnm{Batlle}, \binits{V.M.}},
\bauthor{\bsnm{Montiel}, \binits{J.M.M.}},
\bauthor{\bsnm{Fua}, \binits{P.}},
\bauthor{\bsnm{Tard{\'o}s}, \binits{J.D.}}:
\bctitle{Lightneus: Neural surface reconstruction in endoscopy using illumination decline}.
In: \bbtitle{MICCAI},
pp. \bfpage{502}--\blpage{512}.
\bpublisher{Springer},
\blocation{Cham}
(\byear{2023})
\end{bchapter}
\endbibitem

\bibitem[\protect\citeauthoryear{Batlle et~al.}{2022}]{batlle2022photometric}
\begin{bchapter}
\bauthor{\bsnm{Batlle}, \binits{V.M.}},
\bauthor{\bsnm{Montiel}, \binits{J.M.M.}},
\bauthor{\bsnm{Tard\'os}, \binits{J.D.}}:
\bctitle{Photometric single-view dense {3D} reconstruction in endoscopy}.
In: \bbtitle{IEEE/RSJ Int. Conf. on Intelligent Robots and Systems (IROS)},
pp. \bfpage{4904}--\blpage{4910}
(\byear{2022}).
\doiurl{10.1109/IROS47612.2022.9981742}
\end{bchapter}
\endbibitem

\bibitem[\protect\citeauthoryear{Modrzejewski et~al.}{2020}]{modrzejewski2020light}
\begin{barticle}
\bauthor{\bsnm{Modrzejewski}, \binits{R.}},
\bauthor{\bsnm{Collins}, \binits{T.}},
\bauthor{\bsnm{Hostettler}, \binits{A.}},
\bauthor{\bsnm{Marescaux}, \binits{J.}},
\bauthor{\bsnm{Bartoli}, \binits{A.}}:
\batitle{Light modelling and calibration in laparoscopy}.
\bjtitle{Int. J. Computer Assisted Radiology and Surgery}
\bvolume{15}(\bissue{5}),
\bfpage{859}--\blpage{866}
(\byear{2020})
\end{barticle}
\endbibitem

\bibitem[\protect\citeauthoryear{Deshpande}{2020}]{multiobjtracker_amd2018}
\begin{botherref}
\oauthor{\bsnm{Deshpande}, \binits{A.M.}}:
Multi-object trackers in Python.
GitHub
(2020)
\end{botherref}
\endbibitem

\bibitem[\protect\citeauthoryear{Yang et~al.}{2024}]{depthanything}
\begin{bchapter}
\bauthor{\bsnm{Yang}, \binits{L.}},
\bauthor{\bsnm{Kang}, \binits{B.}},
\bauthor{\bsnm{Huang}, \binits{Z.}},
\bauthor{\bsnm{Xu}, \binits{X.}},
\bauthor{\bsnm{Feng}, \binits{J.}},
\bauthor{\bsnm{Zhao}, \binits{H.}}:
\bctitle{Depth anything: Unleashing the power of large-scale unlabeled data}.
In: \bbtitle{CVPR}
(\byear{2024})
\end{bchapter}
\endbibitem

\bibitem[\protect\citeauthoryear{Sch\"{o}nberger et~al.}{2016}]{schoenberger2016mvs}
\begin{bchapter}
\bauthor{\bsnm{Sch\"{o}nberger}, \binits{J.L.}},
\bauthor{\bsnm{Zheng}, \binits{E.}},
\bauthor{\bsnm{Pollefeys}, \binits{M.}},
\bauthor{\bsnm{Frahm}, \binits{J.-M.}}:
\bctitle{Pixelwise view selection for unstructured multi-view stereo}.
In: \bbtitle{ECCV}
(\byear{2016})
\end{bchapter}
\endbibitem

\bibitem[\protect\citeauthoryear{Sch\"{o}nberger and Frahm}{2016}]{schoenberger2016sfm}
\begin{bchapter}
\bauthor{\bsnm{Sch\"{o}nberger}, \binits{J.L.}},
\bauthor{\bsnm{Frahm}, \binits{J.-M.}}:
\bctitle{Structure-from-motion revisited}.
In: \bbtitle{CVPR}
(\byear{2016})
\end{bchapter}
\endbibitem

\bibitem[\protect\citeauthoryear{Myer~III et~al.}{1994}]{myer1994proposed}
\begin{barticle}
\bauthor{\bsnm{Myer~III}, \binits{C.M.}},
\bauthor{\bsnm{O'Connor}, \binits{D.M.}},
\bauthor{\bsnm{Cotton}, \binits{R.T.}}:
\batitle{Proposed grading system for subglottic stenosis based on endotracheal tube sizes}.
\bjtitle{Annals of Otology, Rhinology \& Laryngology}
\bvolume{103}(\bissue{4}),
\bfpage{319}--\blpage{323}
(\byear{1994})
\end{barticle}
\endbibitem

\bibitem[\protect\citeauthoryear{Ronneberger et~al.}{2015}]{Ronneberger2015}
\begin{bchapter}
\bauthor{\bsnm{Ronneberger}, \binits{O.}},
\bauthor{\bsnm{Fischer}, \binits{P.}},
\bauthor{\bsnm{Brox}, \binits{T.}}:
\bctitle{U-net: Convolutional networks for biomedical image segmentation}.
In: \bbtitle{MICCAI},
pp. \bfpage{234}--\blpage{241}
(\byear{2015}).
\bcomment{Springer}
\end{bchapter}
\endbibitem

\end{thebibliography}

\end{document}